# Adaptive Multiscale Illumination-Invariant Feature Representation for Undersampled Face Recognition

Yang Zhang [a,b], Changhui Hu [a,b,d], Xiaobo Lu[a,b,*]

[a] School of Automation, Southeast University, Nanjing 210096, China

[b] Key Laboratory of Measurement and Control of Complex Systems of Engineering, Ministry of Education, Southeast University, Nanjing 210096, China

[d] School of Automation, Nanjing University of Posts and Telecommunications, Nanjing 210023, China

*Corresponding author, E-mail: xblu2013@126.com (X.B.Lu)

**Abstract-** This paper presents an novel illumination-invariant feature representation approach used to eliminate the varying illumination affection in undersampled face recognition. Firstly, a new illumination level classification technique based on Singular Value Decomposition (SVD) is proposed to judge the illumination level of input image. Secondly, we construct the logarithm edgemaps feature (LEF) based on lambertian model and local near neighbor feature of the face image, applying to local region within multiple scales. Then, the illumination level is referenced to construct the high performance LEF as well realize adaptive fusion for multiple scales LEFs for the face image, performing JLEF-feature. In addition, the constrain operation is used to remove the useless high-frequency interference, disentangling useful facial feature edges and constructing AJLEF-face. Finally, the effects of the our methods and other state-of-the-art algorithms including deep learning methods are tested on Extended Yale B, CMU PIE, AR as well as our Self-build Driver database (SDB). The experimental results demonstrate that the JLEF-feature and AJLEF-face outperform other related approaches for undersampled face recognition under varying illumination.

## 1. INTRODUCTION

With the development of image processing technique, face recognition is commonly used in our daily life, such as in airport, bank and Intelligent Transportation System(ITS) [1-4]. In these applications, frontal face image on ID card or e-passport is served as gallery, meanwhile, the real-time taken face image affected by varying lighting is served as probe [5]. Just like the examples in Fig. 1, the driver images taken by surveillance cameras on highroads are always influenced by serve illumination.

In these situations, undersampled face recognition under severe illumination is a difficult issue to be solved [6].

Aiming at processing the illumination factor in face recognition, a large amount of related approaches appeared. After conclusion, these methods are separated into illumination preprocessing and invariant feature extraction methods. Illumination preprocessing techniques [7-9] take the whole image as process object to reduce the illumination affection. Unfortunately, these methods' satisfied performances are based on strict alignment face images. In addition, they cannot take advantage of the high frequency features, which contain most facial inherent information. Via contrast, illumination invariant feature extraction methods [10-15] are equipped with more efficient illumination processing ability. LOG-DCT [11] aims at realizing the domain changing for face image from spatial domain to frequency domain, extracting the low-frequency DCT coefficients for eliminating. MSLDE [15] extracts the edgemaps of pixels in face image, utilizing Lambertian reflectance model to eliminate illumination effect. However, these methods lack adaptability, only performing well under particular illumination condition.

Based on previous research, generic learning is the conventional solution for undersampled face recognition [1, 2, 16]. Generic learning aims at forming the generic set, consisted of face images which are not belong to the training set, to learn the intra-class variations among the various faces in training set. The adaptive linear regression classification (ALRC) [18] takes use of the k nearest neighbors (kNNs) principle to form intra-class variations. The stacking supervised auto-encoders (SSAE) [19] utilizes the deep supervised auto-encoder to formulate the generic set. However, these formulated generic sets cannot realize effective performances under serious illumination conditions.

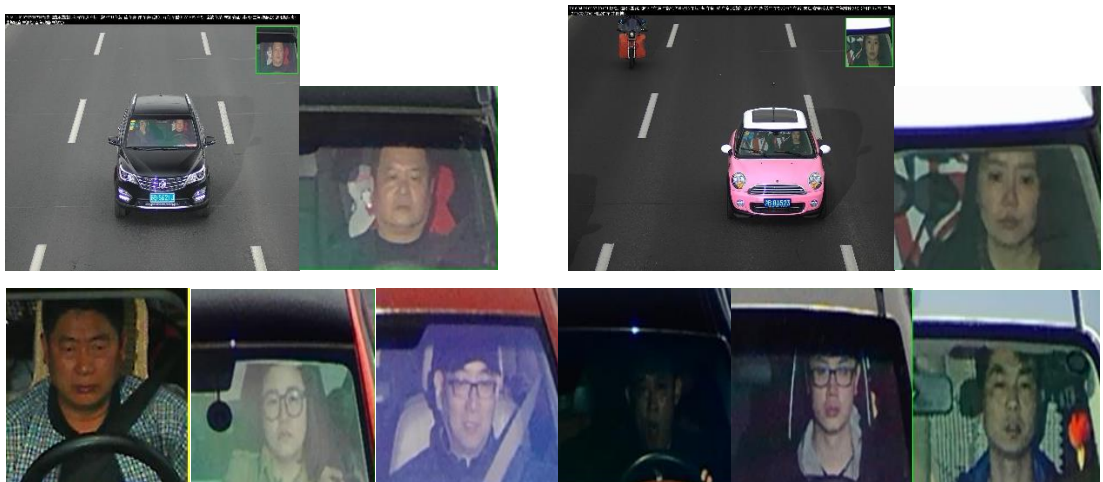

**Fig. 1.** The driver images taken by surveillance cameras on highroads. (Provided by Public Security Department of China)

With the increasingly widespread use of deep learning approaches, deep neural networks have been applied in representation extraction based face recognition [20-26, 47-55]. However, the shortcoming of undersampled face recognition is lack of training samples, which is the fatal flaw for adopting deep learning approach. In addition, to the best of our knowledge, until now, deep learning approaches possess no serious illumination processing ability. The existing deep learning methods, such as multi-view perceptron (MVP) [27], multi-task deep neural network (DNN) [28] as well as disentangled representation learning-generative adversarial network (DR-GAN) [29] are utilized to processing slight illumination and varying pose. In this way, they cannot handle serious illumination changes on face images, such as samples in Extended Yale B [30] and CMU PIE [31] database.

This research proposes a new disentangled illumination-invariant feature representation approach. Firstly, we propose a new illumination level classification approach based on SVD. Then, the illumination intensity is referenced to realize adaptive illumination-invariant feature extraction, constructing JLEF-feature. In addition, the constrain operation is used to remove the useless high-frequency interference from illumination shadows, aiming at retaining useful facial edges, forming AJLEF-face.

Different from the previous researches such as [15, 32, 33], the contributions of this research are as follows:

1) We propose a new illumination level classification approach, judging the corresponding illumination level of input image.

2) We extract JLEF-feature which utilizes the adaptive fusion structure on LEFs based on the corresponding illumination level.

3) The constrain operation is introduced to JLEF-feature to disentangle useful high-frequency facial feature from useless high-frequency interference from illumination shadows, forming the AJLEF-face.

4) This method has been verified on our SDB database, which is constructed in real scene accompanied with varying illumination. Moreover, we take our methods together with other similar state-of-the-art approaches including deep learning researches into comparison on other common datasets, such as Yale B, CMU-PIE and AR databases.

## 2. RELATED WORK

### 2.1. SVD

According to previous study, SVD has been widely used in digital image processing, including feature extraction and noise filter. Andrews et al. [34] utilizes SVD to perform the functions of "low- and high- pass" filter. Later, the fractional order SVD representation (FSVDR) [35] is served as "high-pass" nonlinear filter to eliminate the disturbing noise in face recognition. Demirel et al. [36] and W. Kim et al. [37] held the similar opinion that the corresponding singular matrix can represent the illumination invariant factor of face image. Thus, this research aims at exploring the potential relationship between singular value and illumination level among the face image.

### 2.2. Illumination-Invariant Feature based Illumination Processing

According to the previous study, illumination-invariant feature extraction is known as an effective illumination processing method. Gradient-face[38] and Weber-face[39] were proposed firstly to calculate the illumination insensitive features. The former one is to calculate the ratio between y-gradient with x-gradient, the latter one is to get the ratio between local intensity difference and the constant background. Later, MSLDE is proposed by Lai et al. [34], which utilizes the illumination invariant edges of local area to calculate the illumination invariant feature. However, the common shortage is existing in the above methods, which are all under the assumption that the illumination conditions are similar in local area. Once the illumination shadow in face image varies dramatically, these methods show poor performances. In this paper, we try to construct an adaptive illumination-invariant feature based on illumination intensity and local near neighbor feature, realizing adaptive illumination robust feature extraction.

## 3. THE PROPOSED METHOD

### 3.1. Illumination level classification

This research proposes a new illumination level classification method based on SVD. The classification principle is introduced below.

First of all, the input image is processed by logarithm transformation, showing in Eq.(1).

$$f(x, y) = \ln I(x, y) \tag{1}$$

After logarithm transformation, SVD is introduced to make decomposition on the logarithm face

f(x, y). The processing progress is as following:

$$f(x, y) = UDV^T, D = \begin{pmatrix} d_1 & & & \\ & \cdot & & \\ & & \cdot & \\ & & & \cdot \\ & & & & d_n \end{pmatrix} \quad (2)$$

$$U = [u_1, u_2, ..., u_n] \in R^{m*n} \quad (3)$$

$$V = [v_1, v_2, ..., v_n] \in R^{n*n} \quad (4)$$

Where matrix U and matrix V perform orthogonal essentially. In Eq.(2), singular value $d_1$ to singular value $d_n$ ranks from large to small, in turn. The length and width of input pixel matrix are m and n, respectively.

$$f(x, y) = \sum_{i=1}^{n} d_i u_i v_i^T \quad (5)$$

In Eq.(5) $B_i = u_i v_i^T$ represents ith feature of input face.

$$c_i = Sigmoid(\beta d_i) = \frac{1}{1+e^{-\beta d_i}} \quad (6)$$

Aiming at getting energy coefficient for illumination level (ECIL), ci, to realize classification, we make Sigmoid calculation on all the singular values, di. Under the affection of different β in Sigmoid function, the corresponding ECILs show varying distribution in Fig. 2. We utilize the three images of the same individual from different subset in YaleB to perform the three regular ECIL curves. The intensive degree of ci curves is increasing with the value of β. Based on the experiment result, β=1 leads to the uniform ci curve, classifying the illumination level effectively. Thus, this paper presents a new approach to get the illumination coefficient of corresponding face image, $IL_{coefficient}$, performing in Eq.(7).

$$IL_{coefficient} = \|c_1, c_2, ..., c_n\|_2 \quad (7)$$

This research takes Yale B face database [30], constructed on various illumination conditions, to divide the $IL_{coefficient}$. Thus, the $IL_{coefficient}$s of all the face images in Yale B are calculated. Not surprisingly, the $IL_{coefficient}$ of the face image taken under serious illumination condition is larger than that under normal illumination condition.

Thus, the maximum $IL_{coefficient}$ in Yale B database is defined as Max-$IL_{coefficient}$ and minimum $IL_{coefficient}$ is defined as Max-$IL_{coefficient}$. This research proposes Eq.(8) to calculate the dividing boundary ∇, realizing illumination level classification.

$$\nabla = (Max\text{-}IL_{coefficient} - Min\text{-}IL_{coefficient})/3 \quad (8)$$

In this way, 5 boundary $IL_{level}$s are calculated, as Table 1 shows.

**Table 1**

The classified illumination level

| $IL_{level}$ | $IL_{level}$ Range | |
|---|---|---|
| $IL_{level}0$ | | < Min-IL $_{coefficient}$ |
| $IL_{level}1$ | Min-IL $_{coefficient}$ | Min-IL $_{coefficient}$ $+\nabla$ |
| $IL_{level}2$ | Min-IL $_{coefficient}$ $+\nabla$ | Min-IL $_{coefficient}$ $+2\nabla$ |
| $IL_{level}3$ | Min-ILl $_{coefficient}$ $+2\nabla$ | Max-IL $_{coefficient}$ |
| $IL_{level}4$ | > Max-IL $_{coefficient}$ | |

Based on Table 1, all the images in Yale B can be classified into three illumination conditions. Table 2 performs the processing steps. By convention, Yale B can be used as a benchmark database. Thus, $IL_{level}0$ and $IL_{level}4$ are generated to act as the references for other database's illumination conditions. Fig. 3 shows some example images in various $IL_{level}$ subsets within Yale B database.

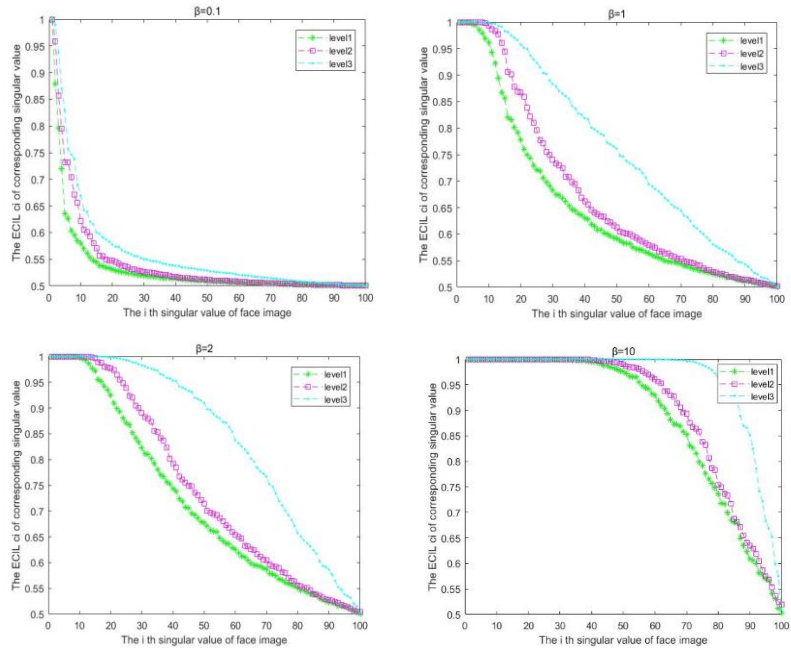

**Fig. 2.** Distributions of ECILs based on different $β$.

**Table 2**

The illumination level ($IL_{level}$) classification steps

Step1 An input image I;

Step2 Logarithmic transformation, performing in Eq.(1);

Step3 SVD decomposition, performing in Eq.(2);

Step4 Calculate the singular value $d_i$;

Step5 Calculate the ECIL $c_i$ of corresponding $d_i$, performing in Eq.(6);

Step6 Calculate the $IL_{coefficient}$, performing in Eq.(7);

Step7 Classify $IL_{level}$ by Table 1.

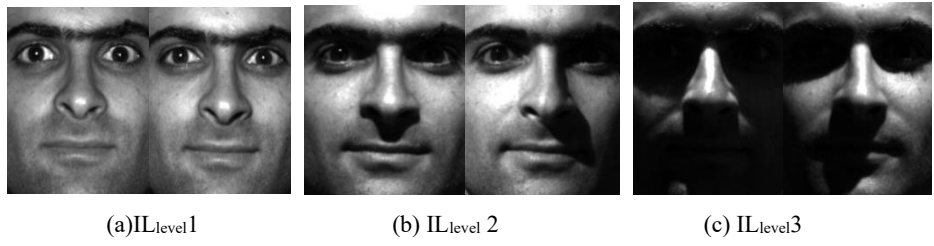

(a) IL$_{level}$1  (b) IL$_{level}$2  (c) IL$_{level}$3

Fig. 3. Example images in various IL$_{level}$ subsets.

**3.2. JLEF feature**

According to previous study, the Lambertian reflectance model [28] is the basic theory for illumination research in face recognition, showing in Eq.(9).

$$I(x,y) = R(x,y) \cdot L(x,y), \ 1 \leq x \leq p, 1 \leq y \leq q \tag{9}$$

Where I is a face image with the gray-scale of p*q. I(x, y) represents the pixel intensity at each pixel point (x, y) among the face image. Based on the previous researches [32, 33], R(x, y) and L(x, y) are the corresponding facial intrinsic feature and illumination component, respectively. Thus, separating L(x, y) from R(x, y) becomes an hot but ill-posed issue.

In order to solve this tricky problem, logarithm transformation is introduced to transform the multiplication model into concise addition model, forming Eq. (10).

$$f(x,y) = \ln I(x,y) = \ln R(x,y) + \ln L(x,y) \tag{10}$$

In Eq(10), f performs logarithm transformation on the original face image I. What should be noticed is that, logarithm transformation can realize enhancement on original image and hold the corresponding intrinsic characteristics.

3.2.1 LEF feature

Weber-face [33] holds the opinion that the illumination component L(x, y) keeps consistent in local pixel area while the facial intrinsic feature R(x, y) changes abruptly. Considering two pixel points $(x_1, y_1)$ and $(x_2, y_2)$ in local area among the face image. The opinion can be expressed as:

$$L(x_1, y_1) \approx L(x_2, y_2) \tag{11}$$

Combining Eq.(10) and Eq. (11), we can get

$$f(x_1, y_1) - f(x_2, y_2) = \ln R(x_1, y_1) + \ln L(x_1, y_1) - (\ln R(x_2, y_2) + \ln L(x_2, y_2))$$
$$= \ln R(x_1, y_1) - \ln R(x_2, y_2) \tag{12}$$

In this research, we formulate the $k_{th}$ LEF feature, which contains all the logarithm difference edgemaps, belonging to various local areas among Π$_k$, as Eq.(13).

$$\text{LEF}_k(x,y) = \sum_{\Pi_k}(f(x,y) - f(x_i, y_i)) = \sum_{\Pi_k}(\ln R(x,y) - \ln R(x_i, y_i)),$$

$$(x_i, y_i) \in Neighbourhood\ ((x,y); size = k) \tag{13}$$

where local areas $\Pi_1 \subset \cdots \subset \Pi_{k-1} \subset \Pi_k$.

Commonly, the illumination robust feature extraction performances, which we defined as discriminative power in this research, of multiple scale ($\Pi_k$ (k=1,2,3,4, $\cdots$)), named as the $k_{th}$ LEF features, are different.

3.2.2 Joint LEF feature

Then, we define the joint LEF feature (JLEF feature) for the local area. JLEF feature can refine the performance of LEF feature by effective fusing all the LEF$_k$s among the local area.

$$\text{JLEF}(x,y) = \sum_{k=1}^{\alpha}(w_g(k) * \text{LEF}_k(x,y))\ ,\ \alpha=1,2,3,4,5\ldots \tag{14}$$

where α is the maximal region size of the local area, which is determined by the corresponding illumination level of the face image in this research. $\omega_g(k)$ is the adaptive fusion weight of LEF$_k$, which is used to adjust the importance and influence of the corresponding LEF$_k$ based on its performance.

3.2.3 Adaptive parameter determination

The Yale B face database [30] with covering a wide range of illumination variations is selected to conduct the parameter determination experiments. According to the illumination dividing technique in Section 2.1, the 2432 face images in Extended Yale B database are redivided into IL$_{level}$ 1-3 subset, accompanying by varying illumination levels from slight to severe, which contain 196, 949, 1203. In IL$_{level}$1 subset, all the face images are under normal illumination condition. For the IL$_{level}$2 subset, the illumination conditions are with minor scale cast shadows. For the IL$_{level}$3 subset, the illumination conditions are with major scale cast shadows.

We exploit the redivided Yale B to estimate the adaptive parameter of JLEF feature and AJLEF-face associated with each IL$_{level}$. Our experiments are composed of two parts. 1)The single training set consists of the first image of each person in subset 1 (i.e. clean training images), and the rest images of Yale B construct the testing set. 2) The first image of each person in subset 5 in the original Yale B database (i.e. unclean training images) forms the single training set, and the Yale B images excluding training ones are designated to test. The nearest neighbor classifier based on Euclidean distance is adopted for the final classification. In this research, all parameter estimations of the proposed methods employ the unclean single training set.

The discriminative power of the $k_{th}$ LEF feature in the redivided subset can be estimated via the experiments. Here, the single training set in the experiment is consisted of unclean training images (the first image of each person in subset 5 in original Yale B database). Fig. 4 shows the recognition results of the multiple scale ($\Pi_k$ (k=1,2,3, ···,10)) $LEF_k$s in each subset. The most important is, Fig. 4 indicates that $LEF_k$s show the best performances when k=5, 4 and 3 for $IL_{level}1$, $IL_{level}2$ and $IL_{level}3$ subsets, respectively.

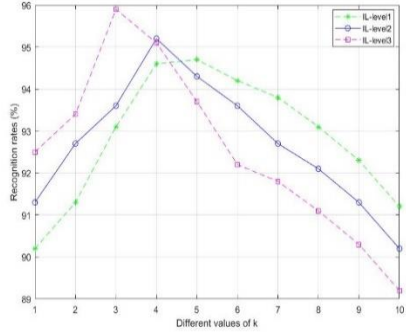

**Fig. 4.** The recognition rates of $LEF_k$ under various k.

From Fig.4, the recognition rates of all the $LEF_k$s in one subset can form the LEF-performance set which reflects the performances of all the $LEF_k$s. Take the $IL_{level}3$ subset in Yale B as example, LEF-performance = [92.5 93.4 95.9 95.1 93.7 92.2 91.8 91.1 90.3 89.2], which just reflects the importance and influence of the corresponding $LEF_k$s for face recognition in $IL_{level}3$ subset. However, due to the small gaps between the parameters among LEF-performance, the performances of effective LNN-features cannot make significant achievement in JLEF feature. Thus, we perform Softmax operation on the parameters in LEF-performance to realize normalization and highlight maximum probability, forming the corresponding $\omega_{normal}$ set.

In addition, according to Fig.4, the $LEF_k$s' performance curves are similar to Gaussian distribution. Aiming at further increasing the gaps between $\omega_{normal}$s, the Gaussian function is utilized to forming Gaussian Weights set $\omega_g$, performing effective feature extraction, showing in Eq.(16).

$$\omega_{normal}(k) = \frac{e^{\text{LEF-performance}(k)}}{\sum_{k=1}^{\alpha} e^{\text{LEF-performance}(k)}} \quad \alpha=1,2,3,4,5\ldots \tag{15}$$

$$\omega_g(k) = \exp\left(-\omega_{normal}(k)^2/2\sigma^2\right) \quad \alpha=1,2,3,4,5\ldots \tag{16}$$

where $\omega_g$ (k) is the $k_{th}$ element in $w$. Through Eq.(15), the more power the $LEF_k$ is, the larger the corresponding weight ($\omega_g$ (k)) of $LEF_k$ is. Eq.(16) enlarges the differences between the parameters among LEF-performance, forming the final weight set $\omega_g$. The same as Fig.4, Fig.5 shows the performance for the JLEF feature under different numbers of $LEF_k$s and various $\sigma^2$ in each subset. Not surprisingly, the values of ks for the best performances of JLEF features in $IL_{level}1$, $IL_{level}2$ and

$IL_{level}3$ subsets are the same to $LEF_k$s'. In addition, for Subset $IL_{level}1$, the best performance is appeared when $\sigma^2=2$. When it comes to $IL_{level}2$ and $IL_{level}3$, $\sigma^2=1$ lead to the best performances.

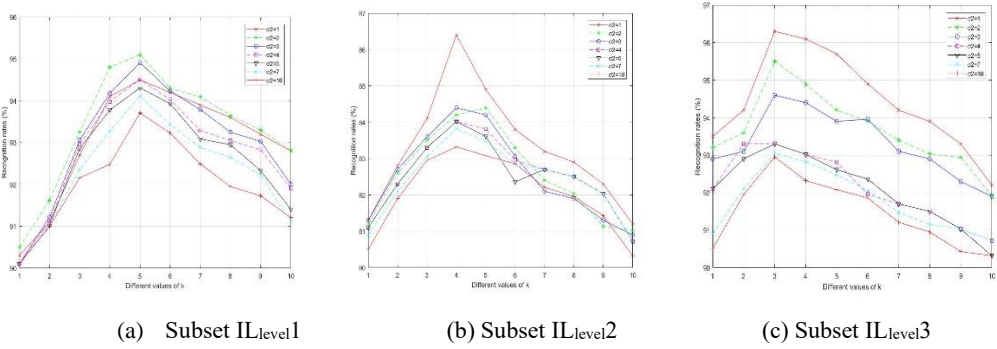

(a) Subset $IL_{level}1$      (b) Subset $IL_{level}2$      (c) Subset $IL_{level}3$

**Fig. 5.** The recognition rates of JLEF feature under various k and $\sigma^2$.

### 3.3. AJLEF-face

However, the JLEF-feature is proposed under the assumption that the illumination components in local area are approximately the same. In fact, noise points and illumination shadows always attack the local area. Thus, the illumination component in Eq.(9) cannot be ignored. Thus, we define the new illumination component in Eq.(17).

$$L(x,y) = (1 + \varepsilon_{i,j})L(x_i, y_j),$$

$$(x_i, y_i) \in Neighbourhood\ ((x,y); size = k) \quad (17)$$

In Eq.(17), $\varepsilon_{i,j}$ reflects the illumination variation between the central pixel point (x, y) with corresponding neighbor point ($x_i$, $y_j$). L(x, y) represents the illumination component of pixel point (x, y). k is the maximal region size of the local area, which is determined by the corresponding illumination level of the face image in this research.

Then, the new calculating model is formed (Eq.(18)).

$$JLEF(x,y) = \sum_{k=1}^{\alpha} \left( \omega_g(k) \sum_{\Pi_k} \left( lnR(x,y) - lnR(x_i, y_i) \right) \right)$$

$$\ln(1 + \varepsilon_{i,j}) \to 0), \alpha = 1,2,3,4,5 \ldots \quad (18)$$

Compared to Gradient-face [32] and Weber-face [33], JLEF-feature is just determined by $ln\ (1 + \varepsilon_{i,j})$, resulting in better processing capacity for illumination changes. The derivation process for Gradient-face [32] and Weber-face [33] and JLEF-feature are shown in Appendix A.

The effect of $ln\ (1 + \varepsilon_{i,j})$ to JLEF-feature is shown in Fig.6, which performs the distribution of gray values of the JLEF-features for the two face images belonging to one individual under various illumination conditions. Based on the assumption of JLEF-feature, these distributions should be consistent, resulting in $\varepsilon_{i,j} \approx 0$. However, Fig.6 shows two different distributions, indicates that the JLEF-feature is inevitably influenced by $ln\ (1 + \varepsilon_{i,j})$, which cannot treated as zero in the model.

In addition, it also validates that the influences of varying lighting can be divided into two parts, one is the high frequency noises including polluted points as well as edges of cast shadows, the other one is the distortions on the high frequency features of face, just like edges of eyes, mouth and nose. The latter influence can also explain the poor performance of deep learning methods on the face dataset influenced by severe illumination variations, due to its requirement for the similar variations of training and validation images.

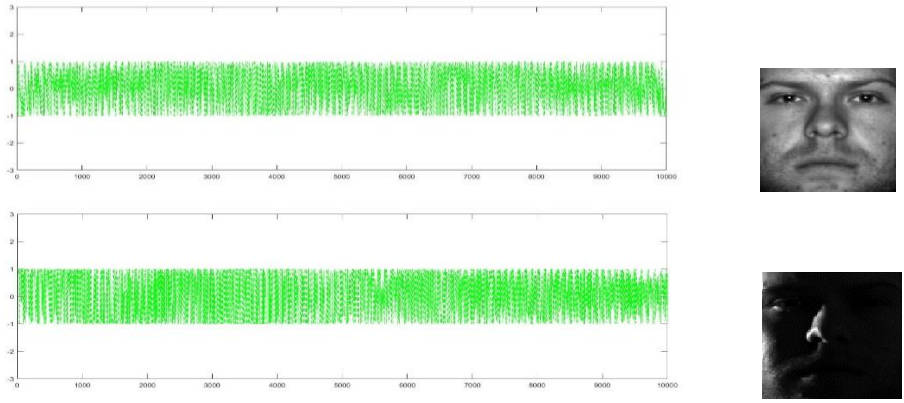

**Fig. 6.** The JLEF-feature ($\alpha$=1) distributions of the face images belonging to the same individual in subset $IL_{level}1$ and $IL_{level}3$.

Above all, the influences of $ln\ (1 + \varepsilon_{i,j})$ to JLEF-feature are the useless high-frequency features produced by varying illumination. However, it is uneasy to distinguish the influence of $ln\ (1 + \varepsilon_{i,j})$ from useful high-frequency facial feature such as eyes, mouth and nose. What is more, the useful facial features are easily polluted by severe illumination changes.

In this research, we adopt the saturation function to constrain the useless high-frequency feature in JLEF-feature. We introduce the sigmoid function to address the high-frequency influence in JLEF-feature, getting the adaptive JLEF face (AJLEF-face).

$$AJLEF - face(x, y) = \frac{1}{1+e^{-\delta JLEF(x,y)}} \qquad (19)$$

In Eq.(19), $\delta$ represents the gain and proportion for the slope of unsaturated part in sigmoid function.

Fig.7 performs the recognition performances of AJLEF-face in different subset under varying values of $\delta$. We can see that AJLEF-face gets the best performance (100%) when $\delta$=3 and 4 in subset $IL_{level}1$, $\delta$=3 in subset $IL_{level}2$ and $\delta$=2 in subset $IL_{level}1$. However, the more larger the value of $\lambda$ is, the more distorted information of JLEF-feature appeared in saturated part. Thus, we choose $\lambda$=3 for AJLEF-face in subset $IL_{level}3$.

In all, according to the above research on Yale B benchmark database, the adaptive values for $\sigma_2$ and $\lambda$ for different illumination level are performed in Table3. The clean images in Yale B may not be as bright as the clean ones in other face databases such as AR [45] or LFW [46], however, the adaptive values are generic for all the clean images, resulting in the same adaptive values in $IL_{level}0$ and $IL_{level}1$. Moreover, Yale B may contain the darkest face image. Hence, $IL_{level}4$ is assigned the same adaptive values as $IL_{level}3$.

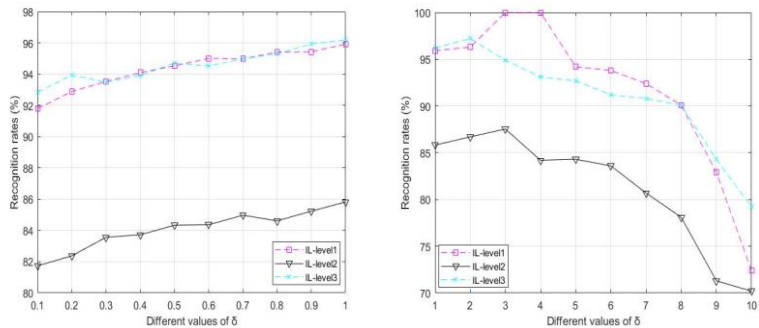

**Fig. 7.** The recognition rates of AJLEF-face under different values of δ.

**Table 3**
The corresponding values for σ2 and λ in various IL-level.

| Subset | k | $\sigma_2$ | δ |
|---|---|---|---|
| $IL_{level}0$ | 5 | 2 | 3 |
| $IL_{level}1$ | 5 | 2 | 3 |
| $IL_{level}2$ | 4 | 1 | 3 |
| $IL_{level}3$ | 3 | 1 | 2 |
| $IL_{level}4$ | 3 | 1 | 2 |

### 3.4. Algorithm framework

Above all, the whole processing procedure is performed in Fig.8. In addition, Fig.9 shows some JLEF features and their corresponding AJLEF-faces. We can see that JLEF features own better visual quantities, but AJLEF-faces can get better recognition rates under varying illumination.

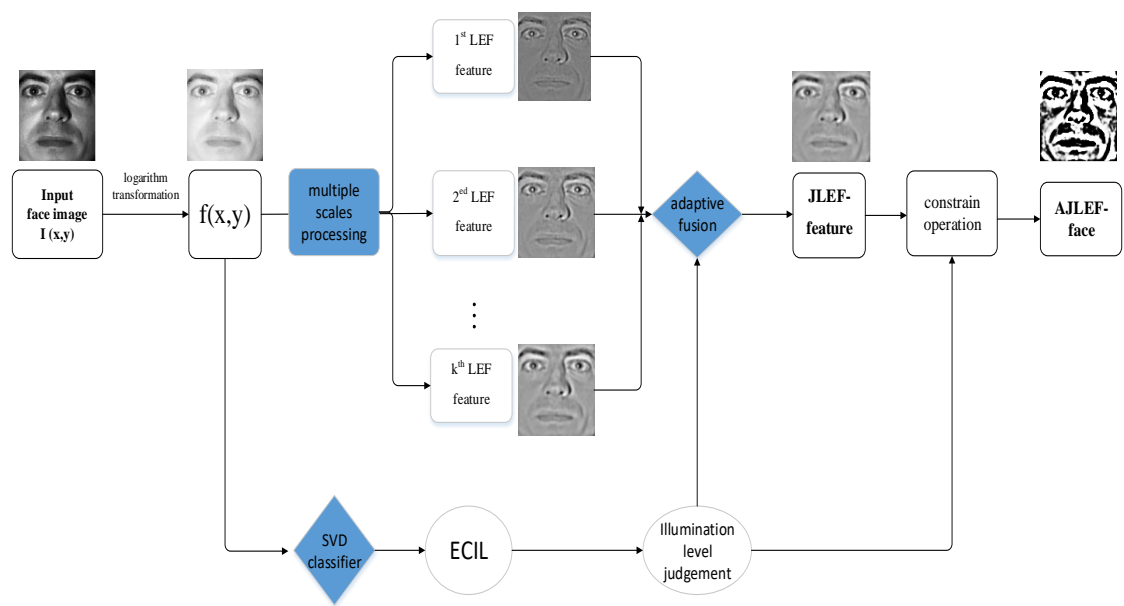

**Fig. 8.** The whole algorithm framework

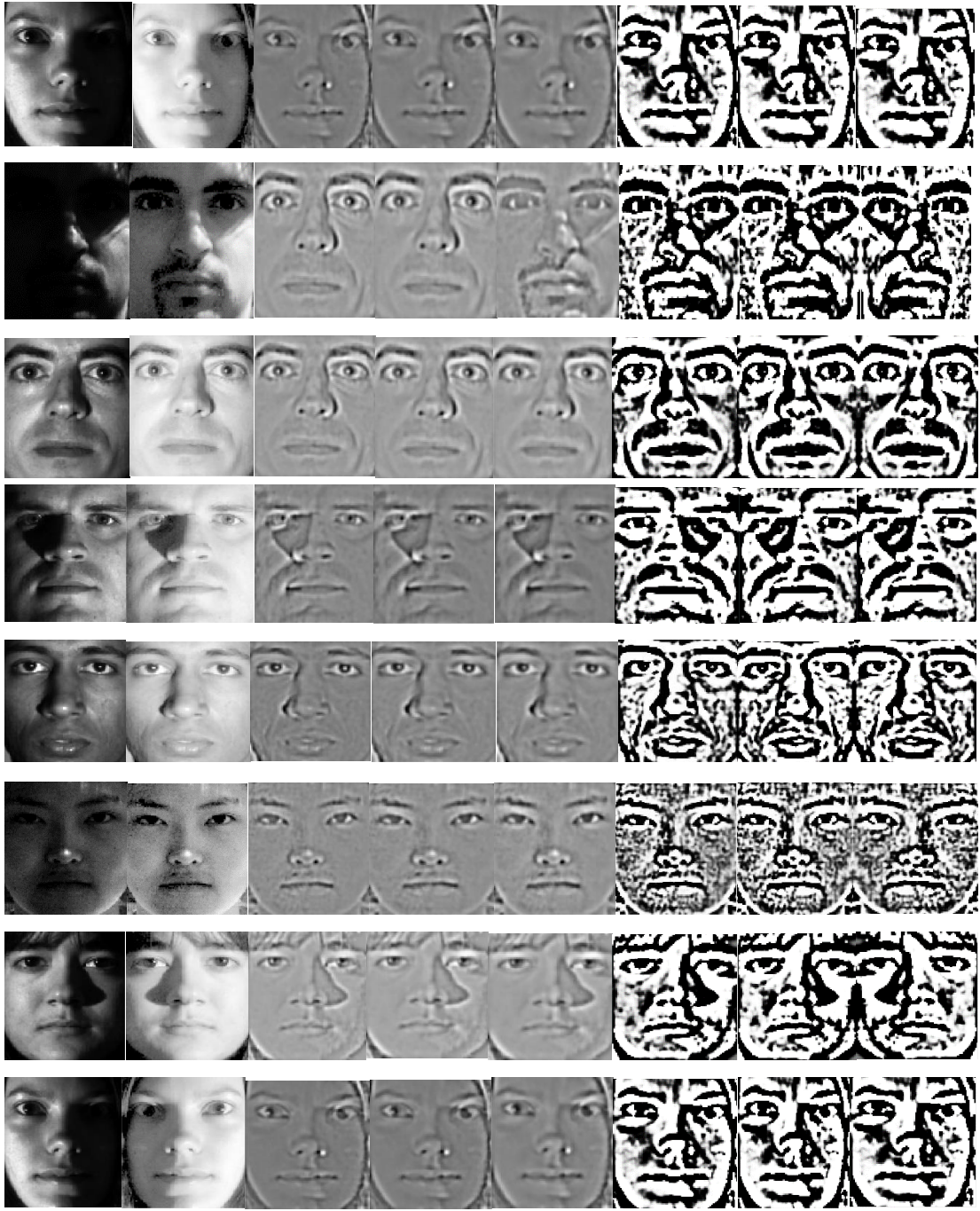

**Fig. 9.** The JLEF features and their corresponding AJLEF-faces. 1st column (left): pixel images; 2nd column: logarithm images; 3rd to 5th columns: JLEF features corresponding to *k*: 3, 4, 5 in Eq. (14); 6th to 8th column: AJLEF-faces corresponding to δ: 1, 2, 3 in Eq. (19).

## 4 EXPERIMENTS

This research conducts JLEF-feature and AJLEF-face methods on Extended Yale B [30], CMU-PIE [31], AR [45], along with our Self-build Driver database to extract the illumination-invariant feature used in face recognition. The state-of-the-art methods including deep learning based methods are taken into comparison.

**4.1. Database Description**

Yale B is a famous public database which is constructed and released by Yale University, containing 38 individuals' face images taken in 9 different poses and 64 varying illumination conditions. This research chooses all the frontal images of the 38 individuals, focusing on the images' illumination issue.

CMU PIE is also a commonly used face database, which includes 41368 face images of 68 individuals based on illumination, expression and varying pose. The P27 subset in CMU PIE, consisted of 1428 images, is taken in 21 different illumination backgrounds, respectively. Thus, it is chose as our experiment sample.

The AR database [33] includes two subsets, containing no less than 4000 face images for 126 individuals. This research utilizes subset 1 as test bed.

To simulate the validity of this research in real traffic scenarios, we construct the SDB database for research. Fig. 10 performs all the cropped face images of one driver. SBD database is consisted of 28 individuals. Each individual has 22 face images taken indoor (12 images) and in car (10 images).

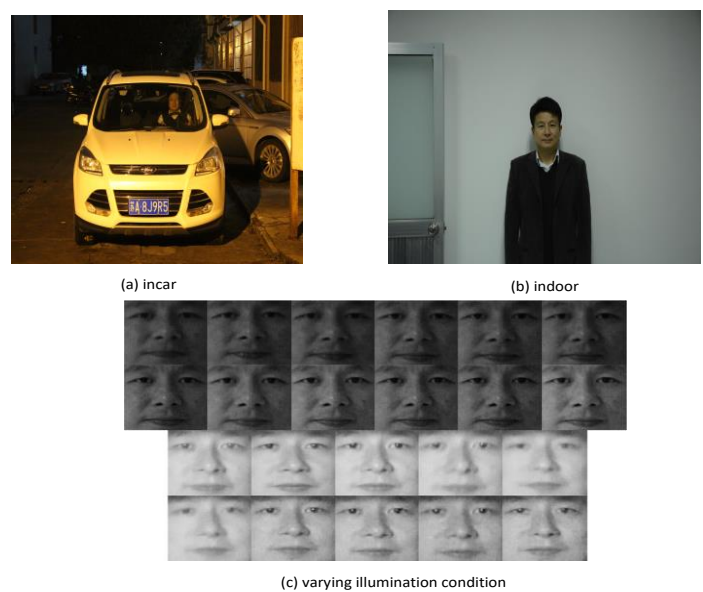

(a) incar  (b) indoor

(c) varying illumination condition

**Fig. 10.** The cropped face images of one individual in SDB database

### 4.2. Experimental Setting

**Baseline.** Average recognition rate (ARR) is proposed in our experiment to evaluate the performances of the compared methods. The gallery image varies from the first image to the last one for one individual, in turn. Then, the other images of the same individual are used as probes. In this way, the test time is the same as the total images' amount for one individual in the dataset. Then, ARR is the average level for all the test results. AAE is a relative fair result compared to other related researches [5-19], which use limited gallery images. Due to the deep learning methods, deep lambertian network (DLN) [42] and SSAE [40], need a large amount of training samples, we choose 1792 face images of the corresponding 28 individuals in Yale B database as training set. Then, the rest images for other 10 persons form the testing set.

**The proposed approach.** JLEF-feature and AJLEF-face.

**Original and LOG.** Original image is the face image with raw pixels. When the logarithm transformation is made on original image, the LOG image is produced. These two style of images undergo no processing, remaining the original facial features used in face recognition. Thus, the extended sparse representation classifier (ESRC) [44] is utilized as classification approach here.

**Deep learning method.** The related deep learning approaches include MADE [40], SSAE [19] as well as VGG [41]. The basic architecture of SSAE is employed from http://www.cs.toronto.edu/_hinton/MatlabForSciencePaper.html. We perform the same experiment set with the original setting in DLN.

**Illumination-invariant feature extraction approaches.** This research takes LOG-DCT [11], LTV [12], MSF [14], Gradient-face [32], MSLDE [15] as well as Weber-face [33] into comparison. According to previous study, $\lambda =0.4$ is set for LTV and $\lambda = 0.1$ is set for MFS. The common parameter $\lambda$ in gaussian kernel filter for Weber-face and MSLDE method is set to 1. The basic architecture codes for Log-DCT, Gradient-face, MSF together with Weber-face were introduced from http://luks.fe.uni-lj.si/sl/osebje/vitomir/ face tools/INFace/index.html. The original code for LTV was referred from http://www.caam.rice.edu/_wy1/ParaMaxFlow/2007/06/binarb-code.html.

### 4.3. Performance Analyzation

The experiment performances of the above methods on Extended Yale B, CMU-PIE, AR and SBD are performed in Table 4 and Table 5. Moreover, Fig. 9 performs part JLEF-features and AJLEF-faces on Yale B and CMU-PIE databases. Based on the above, we can get the conclusion that the performances of JLEF-feature and AJLEF-face are excellent compared to other state-of-the-art methods for extract illumination-invariant representations utilized for face recognition.

(a) Performance on Extended Yale B database.

This research chooses subset 3, 4 and 5 in YaleB dataset, which are constructed under small scale cast shadow, moderate scale cast shadow and large scale cast shadow, respectively, to carry experiment. Even though the face images in Yale B database are influenced by varying illumination conditions, the corresponding facial inherent features still can be extracted. Just like Fig. 9 shows, after logarithm changing, the inherent edges and facial features are revealed. However, previous related methods [5-19] show unsatisfactory extraction performances, especially on the subsets which are influenced by serious illumination.

According to Fig. 11 and Table 4, the JLEF-feature and AJLEF-face proposed by our research perform more efficient feature extraction under varying illumination condition, especially in serious illumination condition. Some previous approaches perform excellent performances under the experiment method that the gallery set contains only one frontal face image under normal illumination condition. However, in our experiment setting, which takes AAE (mentioned in baseline) to judge the whole performance, their performances are lower than JLEF-feature and AJLEF-face. Just like the performances of MSF and MSLDE, which are high in their original researches. When tested under ARR, which shows the average extraction level and relative fair result, their recognition performances cannot keep excellent. On subset 3 and 4 in YaleB, their recognition rates are lower than JLEF-feature and AJLEF-face by gaps of 10% more.

Among all the compared methods, the recognition accuracies of LOG-DCT, LTV, Gradient-face, Weber-face and MSLDE on subset 3 and 4 fall behind subset 5. Thus, we can get the conclusion that they are more suitable for the dataset influenced by large scale cast shadow. Yet, due to lack of training samples, the deep learning techniques, including MADE, SSAE and VGG, cannot achieve good performances, especially on subset 5, which owns relative less samples. Fig. 11 and Table 4 show the recognition performances of all the compared methods in the whole Yale B database.

**Table 4**

The average recognition rates of all the compared techniques in Yale B database

| Approach | Subset3 | Subset4 | Subset5 | Total |
|---|---|---|---|---|
| Original | 33.06% | 19.52% | 15.42% | 20.32% |
| LOG | 40.63% | 31.24% | 32.16% | 22.42% |

| Method | | | | |
|---|---|---|---|---|
| LOG-DCT [11] | 77.44% | 68.24% | 93.22% | 77.95% |
| LTV [12] | 76.62% | 66.43% | 81.97% | 67.63% |
| MSF [14] | 84.82% | 69.46% | 83.99% | 69.85% |
| Gradient-face [32] | 78.22% | 73.51% | 91.02% | 80.75% |
| Weber-face [33] | 79.42% | 76.36% | 95.34% | 83.95% |
| MSLDE [15] | 71.42% | 73.98% | 95.44% | 82.08% |
| MADE [40] | 43.41% | 29.26% | 20.31% | 29.12% |
| SSAE [19] | 44.23% | 28.18% | 21.76% | 30.17% |
| VGG [41] | 36.64% | 23.64% | 17.08% | 21.73% |
| JLEF-feature | **98.01%** | **86.39%** | **96.04%** | **86.29%** |
| AJLEF-face | **98.66%** | **87.54%** | **97.22%** | **88.61%** |

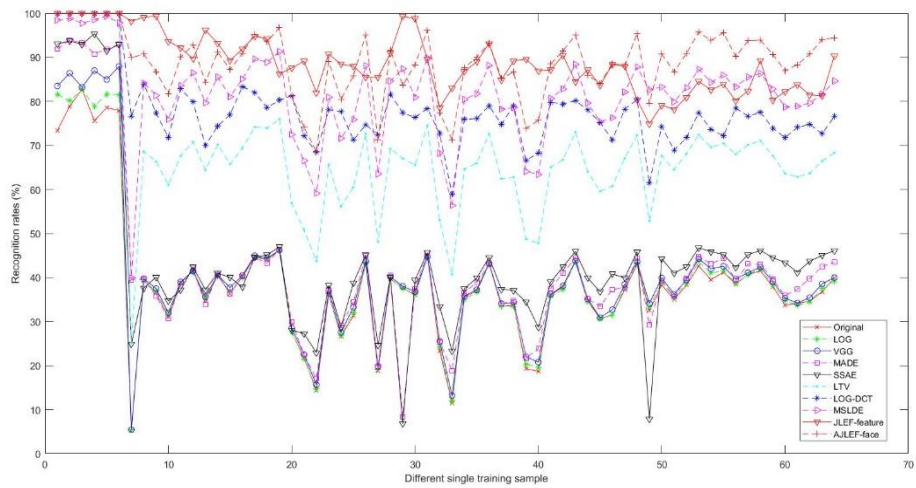

**Fig. 11.** Average recognition accuracy of the compared techniques in the whole Yale B database.

(b) Results on CMU PIE, AR and SDB databases.

1) Results on CMU PIE. From Table 5, the JLEF-feature and AJLEF-face still outperform other comparative methods on CMU PIE database, ranking 1$^{st}$ and 2$^{ed}$. Due to the fact that the illumination condition in CMU PIE database is not so various as it in Yale B. Thus, the differences of the compared methods' performances on CMU PIE database are not so remarkable. Not surprisingly, deep learning methods, including MADE, SSAE and VGG, still cannot achieve satisfied performances, owning to less training samples under similar illumination condition as well as unsupervised learning setting. Compared to other methods, the JLEF-feature and AJLEF-face contain the restriction factor for illumination component and enhancement factor for facial component to ensure the identity invariance. Thus, the superiority of JLEF-feature and AJLEF-face for undersampled face recognition on CMU PIE database is significant. Fig. 12 indicates the average performances of the related methods in the whole CMU PIE database.

2) Results on AR and SBD. According to Table 5, AJLEF-face shows best recognition accuracy on AR, owning the excellent illumination robust feature extraction ability. According to Fig. 10, the illumination condition in SBD are moderate. Not surprisingly, JLEF-feature and AJLEF-face outperform other compared methods on SBD, whereas the margins are not as large as on other face datasets.

**Table 5**

The average recognition rates (%) of the above methods in CMU PIE, AR and SBD databases

| Approach | CMU PIE | AR | SBD |
|---|---|---|---|
| Original | 29.81% | 53.51% | 32.38% |
| LOG | 30.36% | 67.12% | 35.30% |
| LOG-DCT[11] | 70.65% | 56.11% | 51.79% |
| LTV [12] | 71.18% | 60.12% | 53.75% |
| Gradient-face [32] | 87.66% | 57.39% | 60.42% |
| Weber-face [33] | 88.90% | 60.42% | 63.38% |
| MSLDE [15] | 90.11% | 63.28% | 68.94% |
| MFS [14] | 59.12% | 43.71% | 50.91% |
| SSAE [19] | 68.29% | 50.94% | 42.26% |
| VGG [41] | 50.03% | 47.07% | 49.80% |
| JLEF-feature | 90.64% | 68.88% | 72.56% |
| AJLEF-face | 91.33% | 71.02% | 74.72% |

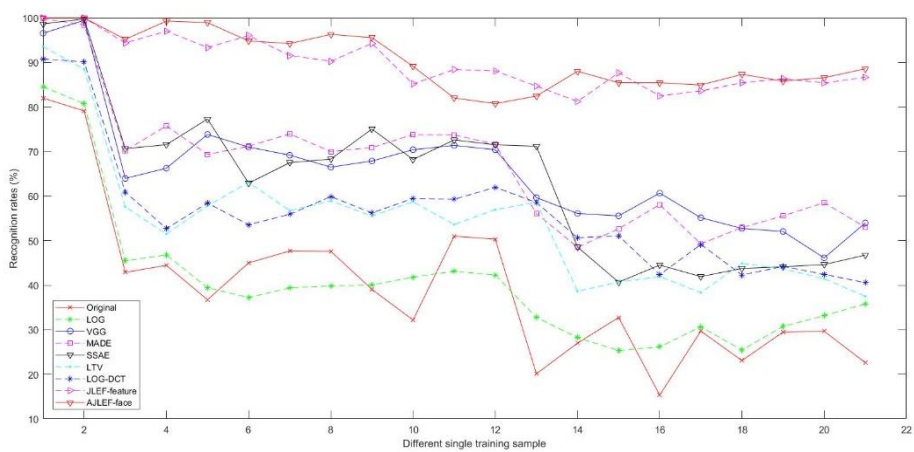

**Fig. 12.** The average performances of the related methods in the whole CMU PIE database.

## 5. Conclusion

In this paper, we propose JLEF-feature and AJLEF-face methods based on illumination level classification, extracting the illumination-invariant feature as well preserving inherent facial information. Comparative trials on Yale B, CMU-PIE, AR as well as SBD databases show that the proposed methods own better performances when compared to other feature extraction techniques, including LOG-DCT, MSF, Gradient-face, LTV, Weber-face and MSLDE, as well as related deep learning approaches, including MADE, VGG and SSAE. It indicates that JLEF-feature and AJLEF-face outperform other state-of-the-art methods used in undersampled face recognition affected by varying illumination. In the next research, we will improve this work to make it not only equipped with illumination robustness but also pose and occlusion robustness by adjusting the algorithm structure.

**Acknowledgments**

This work was supported by the National Natural Science Foundation of China (No.61871123 &No.61802203), Key Research and Development Program in Jiangsu Province (No.BE2016739), Natural Science Foundation of Jiangsu Province (No.BK20180311) and a Project Funded by the Priority Academic Program Development of Jiangsu Higher Education Institutions.

**Declarations of interest**

None.

**Appendix A**

The influences of illumination difference $\varepsilon_{i,j}$ to Weber-face, Gradient-face and JLEF-feature are as below.

$$\text{Weber} - \text{face} = \arctan\left(\sum_{\Pi_k} \frac{I(x,y) - I(x_i, y_j)}{I(x,y)}\right)$$

$$= \arctan\left(\sum_{\Pi_k} \frac{R(x,y)L(x,y) - R(x_i, y_j)L(x_i, y_j)}{R(x,y)L(x,y)}\right)$$

$$= \arctan\left(\sum_{\Pi_k} \frac{R(x,y)L(x,y) - R(x_i, y_j)(1 + \varepsilon_{i,j})L(x,y)}{R(x,y)L(x,y)}\right)$$

$$= \arctan\left(\sum_{\Pi_k} \frac{R(x,y) - R(x_i, y_j)(1 + \varepsilon_{i,j})}{R(x,y)}\right)$$

$$= \arctan\left(\sum_{\Pi_k} \frac{R(x,y) - R(x_i, y_j)}{R(x,y)} - \frac{R(x_i, y_j)}{R(x,y)}\varepsilon_{i,j}\right)$$

$$= \arctan\left(\sum_{\Pi_k} \frac{R(x,y) - R(x_i, y_j)}{R(x,y)}\right), \frac{R(x_i, y_j)}{R(x,y)}\varepsilon_{i,j} \to 0$$

$$\text{Gradient} - \text{face} = \arctan\left(\frac{\partial_y I(x,y)}{\partial_x I(x,y)}\right)$$

$$= \arctan\left(\frac{\partial_y R(x,y)L(x,y)}{\partial_x R(x,y)L(x,y)}\right)$$

$$= \arctan\left(\frac{\frac{R(x,y+\Delta yj)L(x,y+\Delta yj) - R(x,y)L(x,y)}{\Delta yj}}{\frac{R(x+\Delta xi,y)L(x+\Delta xi,y) - R(x,y)L(x,y)}{\Delta xi}}\right)$$

$$= \arctan\left(\frac{\frac{R(x,y+\Delta yj)(1+\varepsilon_{i,j})L(x,y) - R(x,y)L(x,y)}{\Delta yj}}{\frac{R(x+\Delta xi,y)(1+\varepsilon_{i,j})L(x,y) - R(x,y)L(x,y)}{\Delta xi}}\right)$$

$$= \arctan\left(\frac{\frac{R(x,y+\Delta yj) - R(x,y) + R(x,y+\Delta yj)\varepsilon_{i,j}}{\Delta yj}}{\frac{R(x+\Delta xi,y) - R(x,y) + R(x+\Delta xi,y)\varepsilon_{i,j}}{\Delta xi}}\right)$$

$$= \arctan\left(\frac{\frac{R(x,y+\Delta yj) - R(x,y)}{\Delta yj} + \frac{R(x,y+\Delta yj)\varepsilon_{i,j}}{\Delta yj}}{\frac{R(x+\Delta xi,y) - R(x,y)}{\Delta xi} + \frac{R(x+\Delta xi,y)\varepsilon_{i,j}}{\Delta xi}}\right)$$

$$= \arctan\left(\frac{\partial_y R(x,y) + \frac{R(x,y+\Delta yj)\varepsilon_{i,j}}{\Delta yj}}{\partial_x R(x,y) + \frac{R(x+\Delta xi,y)\varepsilon_{i,j}}{\Delta xi}}\right)$$

$$= \arctan\left(\frac{\partial_y R(x,y)}{\partial_x R(x,y)}\right), \frac{R(x,y+\Delta yj)\varepsilon_{i,j}}{\Delta yj} \to 0, \frac{R(x+\Delta xi,y)\varepsilon_{i,j}}{\Delta xi} \to 0$$

$$\text{JLEF} - \text{feature} = \sum_{k=1}^{\alpha}\left(\omega_g(k) * \text{LEF}_k(x,y)\right)$$

$$= \sum_{k=1}^{\alpha}\left(\omega_g(k) \sum_{\Pi_k}(f(x,y) - f(x_i,y_i))\right)$$

$$= \sum_{k=1}^{\alpha}\left(\omega_g(k) \sum_{\Pi_k}\left((lnR(x,y) + lnL(x,y)) - (lnR(x_i,y_i) + lnL(x_i,y_i))\right)\right)$$

$$= \sum_{k=1}^{\alpha}\left(\omega_g(k) \sum_{\Pi_k}\left((lnR(x,y) + lnL(x,y)) - (lnR(x_i,y_i) + \ln(1+\varepsilon_{i,j}) + lnL(x,y))\right)\right)$$

$$= \sum_{k=1}^{\alpha}\left(\omega_g(k) \sum_{\Pi_k}\left(lnR(x,y) - lnR(x_i,y_i) - \ln(1+\varepsilon_{i,j})\right)\right)$$

$$= \sum_{k=1}^{\alpha}\left(\omega_g(k) \sum_{\Pi_k}\left(lnR(x,y) - lnR(x_i,y_i)\right)\right), \ln(1+\varepsilon_{i,j}) \to 0), \alpha = 1,2,3,4,5\ldots$$

**Appendix B**

Some processed results of the compared methods: original, logarithm processing, SQI, weber-face, gradient-face, MSLDE, JLEF-feature and AJLEF-face.

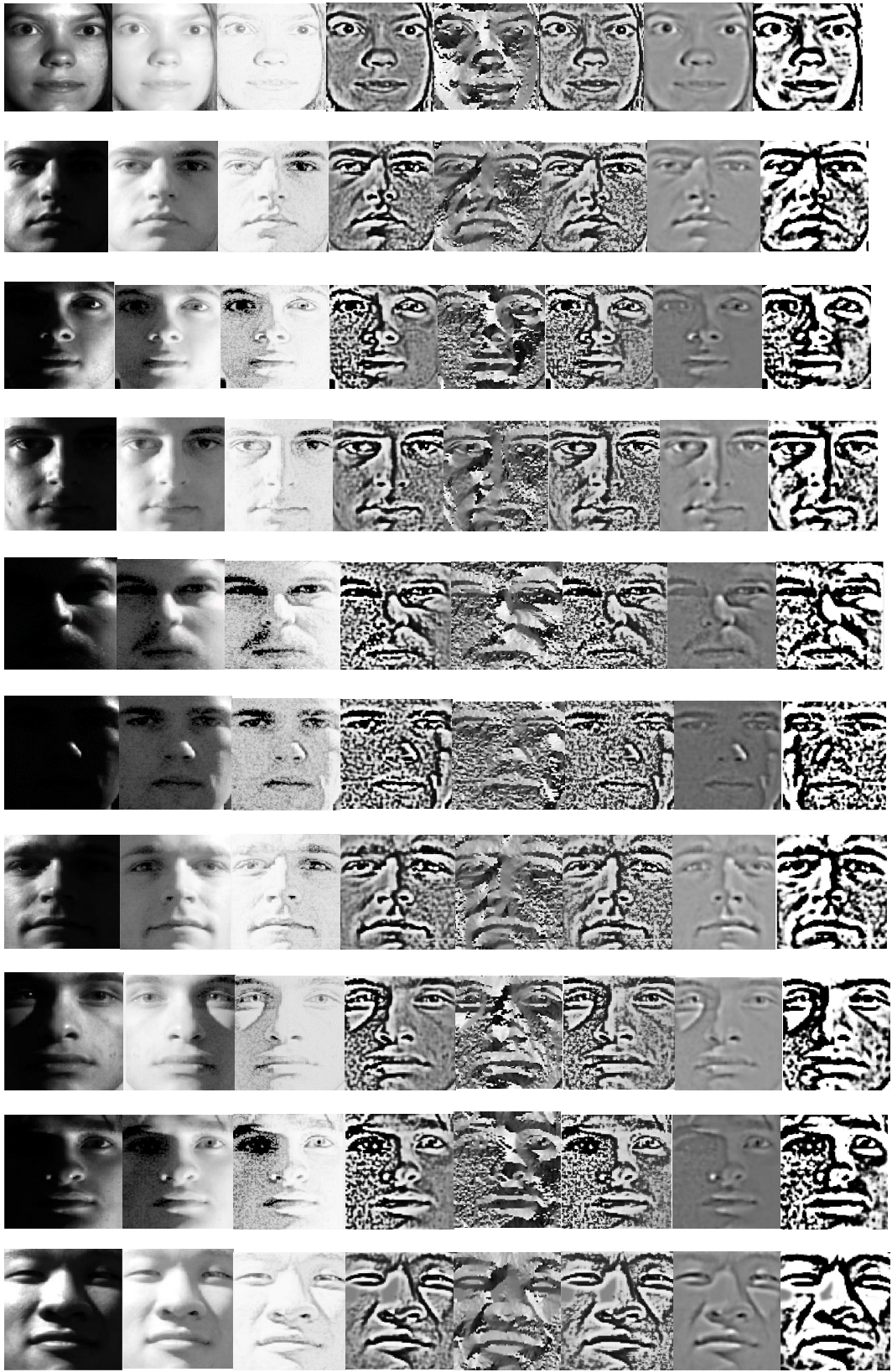